%% file: main.tex
\documentclass[10pt,twocolumn,letterpaper]{article}

\usepackage{cvpr}              

\input{preamble}

\definecolor{cvprblue}{rgb}{0.21,0.49,0.74}
\usepackage[pagebackref,breaklinks,colorlinks,allcolors=cvprblue]{hyperref}

\def\paperID{*****} 
\def\confName{CVPR}
\def\confYear{2026}

\title{Structured Relational Reasoning for Group Activity Assessment
}
\author{
Thinesh Thiyakesan Ponbagavathi\textsuperscript{1,2}\thanks{Equal contribution.} \quad
Chengzheng Yang\textsuperscript{1}\footnotemark[1] \quad
Alina Roitberg\textsuperscript{2} \\
\textsuperscript{1}\small University of Stuttgart, Germany \\
\textsuperscript{2}\small University of Hildesheim, Germany
}

\begin{document}
\maketitle
\input{sec/0_abstract}    
\input{sec/1_intro}

\input{sec/2_related_work}

\input{sec/3_method}
\input{sec/4_experiments}

\input{sec/5_conclusion}
{
    \small
    \bibliographystyle{ieeenat_fullname}
    \bibliography{main}
}


\end{document}

%% file: preamble.tex
\usepackage{pifont}



%% file: sec/0_abstract.tex
\begin{abstract}
Group Activity Detection (GAD) involves recognizing social groups and their collective behaviors in videos.  
 Vision Foundation Models (VFMs), like DINOv2, offer excellent features but are pretrained on object-centric data. We find that naively substituting them into existing GAD pipelines actually degrades performance, exposing structured group-aware decoding as the true bottleneck.

We introduce ProGraD, a structured relational-reasoning framework for GAD built on top of frozen VFMs.  At its core is a lightweight two-layer GroupContext Transformer that explicitly models actor–group associations and aggregates global context to infer collective behavior.  Learnable group prompts serve as a minimal conditioning mechanism to guide the frozen backbone toward socially relevant representations, while the relational decoder performs the core reasoning over actors and groups.  This design jointly infers group locations, memberships, and activities in a single pass using only $\sim$10M trainable parameters - less than half of prior methods.
On the Café benchmark with multiple concurrent social groups, ProGraD improves the state-of-the-art by 6.5\% Group mAP$@$1.0 and 8.2\% Group mAP$@$0.5. On Social-CAD, it achieves state-of-the-art social and membership accuracy. ProGraD further produces interpretable attention maps that provide insights into actor–group reasoning.
\end{abstract}

%% file: sec/1_intro.tex
\begin{figure}[t]
\centering
\includegraphics[width=1\columnwidth]{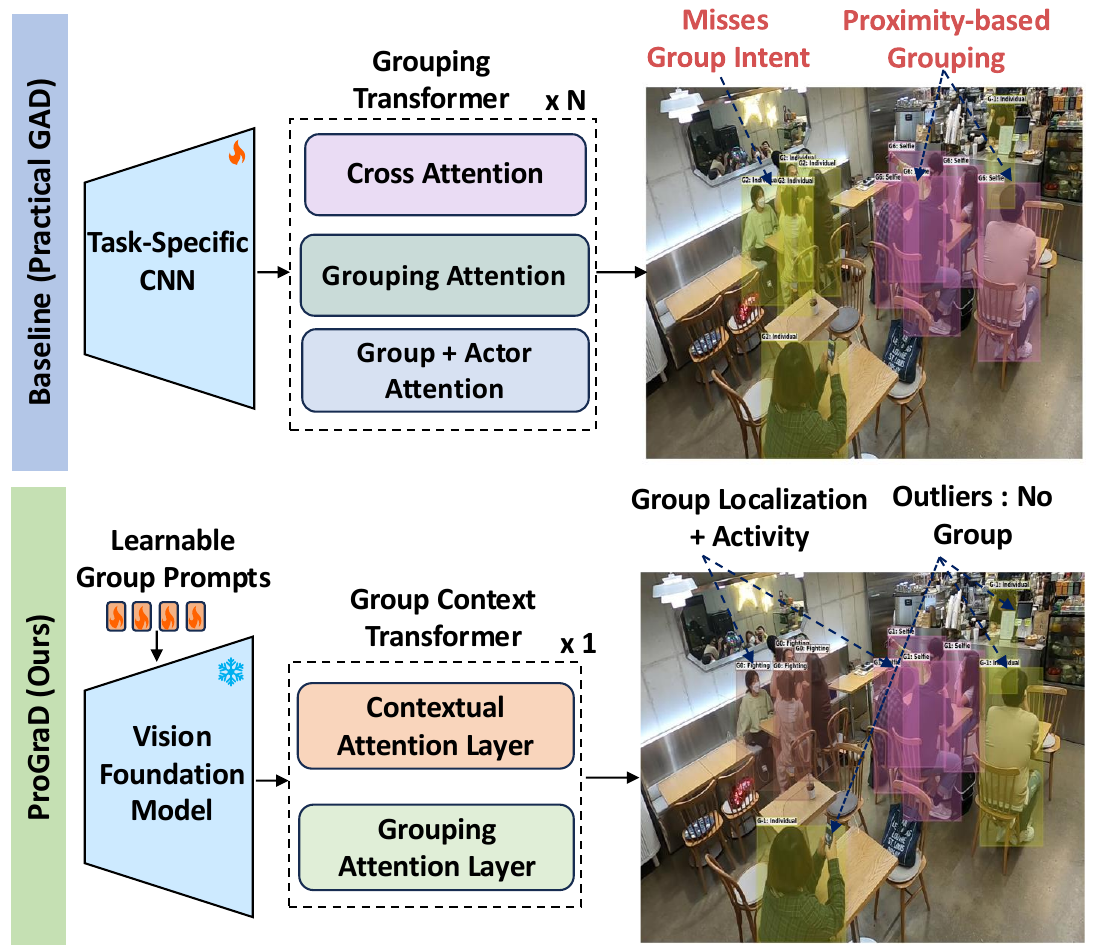} 
\caption{\textbf{ProGraD for Group Activity Detection.} 
Existing GAD methods (top) rely on fully fine-tuned backbones and complex decoders that often fail to capture social intent. ProGraD (bottom) employs a lightweight GroupContext Transformer to explicitly model actor–group relations, with learnable group prompts providing minimal conditioning. This design enables joint group localization, membership inference, and activity recognition in a single pass, yielding consistent improvements over prior methods.}
\vspace{-2em}
\label{fig:Teaser}
\end{figure} 
\section{Introduction}
\label{sec:intro}

Group Activity Detection (GAD) \cite{kim2024cafe,tamura2022huntinggroupcluestransformers,ehsanpour2020joint,ehsanpour2022jrdb} has broad applications across sports analytics, surveillance, classroom monitoring, human-robot collaboration, and group skill assessment.  In these settings, evaluating skilled performance depends not only on recognizing what activity is being performed, but on understanding how individuals coordinate, assume roles, and contribute to the collective execution of a group action. Meaningful assessment and feedback generation therefore require reasoning over group structure - identifying who is acting together, how they interact, and where these interactions occur.

GAD formalizes this challenge by extending conventional group activity recognition \cite{promptGAR,zhou2022composer,graph_GAR} to localize multiple groups within a scene and classify their collective activities. 
The task involves predicting \textit{who} is acting, \textit{with whom}, and \textit{where} these groups are located. This setting is particularly challenging because group membership must be inferred dynamically while modeling both individual and group-level behaviors under overlapping interactions, occlusions, and noisy observations.

Most existing methods \cite{kim2024cafe, tamura2022huntinggroupcluestransformers} rely on fully fine-tuned  CNN backbones with complex spatiotemporal group reasoning modules, resulting in many trainable parameters and relying on heuristic group formation strategies. Recent approaches \cite{gad_vlm,gad_vlm1} leverage pretrained representations, including multimodal foundation models, but retain deep, multi-stage decoders, increasing architectural and training complexity. Vision Foundation Models (VFMs)~\cite{bommasani2021opportunities} have demonstrated strong generalization across downstream tasks and are commonly adapted using lightweight \textit{probing} \cite{linear_prob_simclr} strategies, where only a small set of parameters is trained on top of frozen backbones. However, their potential for group activity understanding remains largely underexplored, as it requires explicit modeling of actor–group relations not captured by standard GAD frameworks.




We find that naively replacing CNN backbones with frozen VFMs in existing GAD pipelines \cite{kim2024cafe} degrades performance. Group mAP$@$1.0 drops from 10.85 to 9.46 when replacing the ResNet-18 backbone in Practical GAD with DINOv2. 
Notably, even approaches incorporating strong visual–language models observe only marginal improvements despite their enhanced representational capacity \cite{gad_vlm, gad_vlm1}. 
 This indicates that the primary bottleneck lies not in feature quality, but in the absence of decoding mechanisms for actor–group relations. Prior analyses \cite{non_linear_dc_probes, non_linear_probing_classifer} further show that expressive probing classifiers can learn signals not encoded in the backbone. This motivates building lightweight, structured decoders that align with, rather than override, the representational capacity of frozen VFMs.

Motivated by this, we propose \textbf{ProGraD}, a structured relational decoding framework that adapts frozen VFMs for GAD. At its core is a lightweight two-layer \textbf{GroupContext Transformer (GCT)} that explicitly models actor–group associations and aggregates global context to infer collective behavior. To expose socially relevant cues without expressive probes, we employ learnable group prompts as a minimal conditioning mechanism, placing the primary reasoning burden on the relational decoder. We evaluate on two benchmarks: Café \cite{kim2024cafe}, featuring multiple concurrent social groups, and Social-CAD \cite{ehsanpour2020joint}, where most groups are singletons. ProGraD outperforms prior methods on both - with gains of 6.5\% Group mAP@1.0 and 8.2\% Group mAP@0.5 on Café, using only $\sim$10M trainable parameters, less than half of prior methods. These gains hold in both frozen and fully fine-tuned settings, confirming they stem from structured relational decoding rather than backbone scale.

To summarize, our key contributions are: (1) We demonstrate that naively substituting VFMs into existing GAD decoders degrades performance, identifying group-aware decoding, not feature quality, as the key bottleneck.
 (2) We introduce the \textbf{GroupContext Transformer (GCT)}, a lightweight two-layer relational decoder that explicitly separates actor-group formation from global context aggregation.
(3) We present \textbf{ProGraD}, a framework integrating GCT with frozen VFMs via minimal PEFT conditioning, achieving state-of-the-art performance on Café and Social-CAD in both frozen and fully fine-tuned settings with ~10M trainable parameters - less than half of prior methods.

%% file: sec/2_related_work.tex
\section{Related Work}
\label{sec:formatting}

\noindent\textbf{Group Activity Detection} 
(GAD) extends traditional group activity recognition \cite{ehsanpour2022jrdb, tamura2022huntinggroupcluestransformers,zhou2022composer} by requiring models to localize multiple groups within a scene and classify each group’s activity. This involves both group membership prediction and group-level activity classification. Early approaches \cite{ehsanpour2022jrdb,han2022panoramic, graph_GAR} employed graph neural networks (GNNs) with spectral clustering to model actor relationships, but relied on non-task-specific clustering algorithms, leading to slow inference and limited adaptability in dynamic scenes.
Recent transformer-based methods address these shortcomings by unifying localization and reasoning. HGC \cite{tamura2022huntinggroupcluestransformers} performs group localization in the 2D coordinate space by applying Deformable DETR \cite{detr} to match actors to predicted group centers using spatial cues. In contrast, Practical GAD \cite{kim2024cafe}  introduces learnable group tokens that enable matching in the embedding space, allowing the model to reason over semantic similarity rather than spatial proximity. 
Recent work \cite{gad_vlm} incorporates pretrained vision–language foundation models to enrich visual and semantic representations, but typically retains decoding architectures originally designed for CNN backbones.
In contrast, ProGraD adopts a lightweight two-layer relational decoder that explicitly reasons over features from a frozen VFM backbone and uses minimal PEFT-based conditioning \cite{jia2022vpt,adaptformer} to capture actor–group semantics.

\begin{figure*}[t]
\centering
\includegraphics[width=1\textwidth]{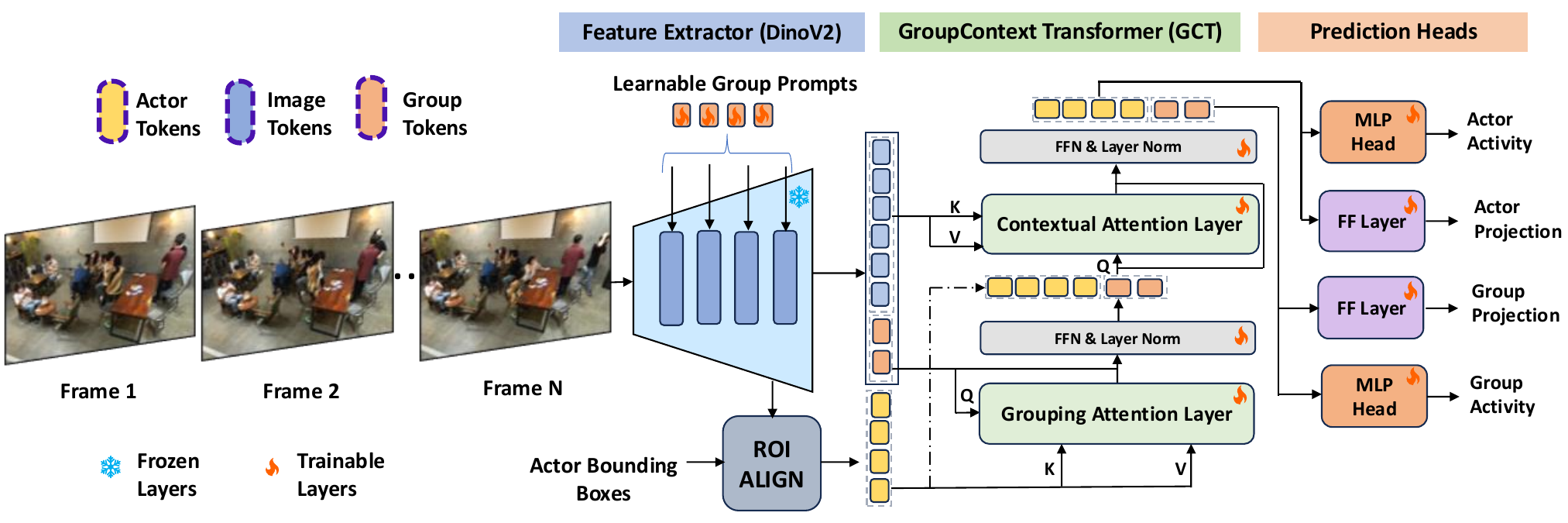} 
\caption{\textbf{Overview of ProGraD}. ProGraD consists of three key components: (1) a frozen feature extractor enhanced with learnable group prompts for group-aware feature extraction; (2) a GroupContext Transformer (GCT) that separates actor-group formation from global context aggregation; and (3) lightweight prediction heads for group activity, actor activity, and membership inference.}
\label{fig:Architecture}
\vspace{-1em}
\end{figure*}

\noindent\textbf{Lightweight Adaptation of Vision Foundation Models}
Vision Foundation Models \cite{CLIP,dinov2,mae} are typically adapted via linear probing \cite{linear_prob_representationlearningcontrastivepredictive,linear_prob_selfsupervisedvisualrepresentation} or attentive strategies \cite{set_transformer, coca} for classification-based tasks. More expressive adaptation is achieved through Parameter-Efficient Fine-Tuning (PEFT), using adapters \cite{St_adaptor, aim} or prompts \cite{jia2022vpt,VitaCLIP} to inject task specificity into frozen backbones. However, these approaches are designed for per-instance recognition and do not address structured relational reasoning over multiple interacting agents. 
 In the context of group activity recognition, PromptGAR \cite{promptGAR} unifies visual input annotations - bounding boxes, keypoints, and area definitions - as point prompts for input flexibility, but addresses the simpler GAR setting, classifying a manually selected group without localizing multiple concurrent groups. In contrast, ProGraD uses learnable soft prompts or adapters as minimal backbone conditioning, with the GroupContext Transformer performing actor-group reasoning across concurrent groups.

%% file: sec/3_method.tex
\section{Methodology}

We propose \textbf{ProGraD}, a structured relational decoding framework for GAD that adapts frozen VFMs following the Parameter-Efficient Fine-Tuning (PEFT) \cite{jia2022vpt,VFT,aim,St_adaptor} paradigm. 
The central component is the \textbf{GroupContext Transformer (GCT)}, a lightweight two-layer relational decoder that explicitly models actor–group associations and aggregates global scene context to infer collective behavior. To expose socially relevant semantics from the frozen backbone, we employ lightweight PEFT-based conditioning as a minimal mechanism, placing the primary reasoning burden on the GCT.
Together, this design enables joint prediction of group membership, group activity, and individual actions, while naturally handling varying numbers of groups. 
Figure~\ref{fig:Architecture} provides an overview of the architecture.

\subsection{Problem definition and notation}

Given a video $\mathbf{X} = \{X_1, X_2, \ldots, X_T\}$, our goal is to detect social groups and classify their collective activities. Each frame $X_t$ is processed through a VFM, such as DinoV2 \cite{dinov2}, yielding patch-level tokens $I_t \in \mathbb{R}^{H_fW_f \times D}$, where $H_fW_f$ is the number of patch tokens and $D$ is the feature dimension. These frame features are stacked to obtain image features $\mathbf{I} \in \mathbb{R}^{H_fW_f \times T \times D}$.  We additionally extract actor tokens $\mathbf{A} = \{a_1, a_2, \ldots, a_M\} \in \mathbb{R}^{M \times T \times D}$ using ROI Align guided by actor bounding box annotations, where $M$ denotes the total number of actor instances per frame. 
To enable structured group-level reasoning, we introduce a set of learnable group prompts $\mathbf{G_{prompt}} = \{g_1, g_2, \ldots, g_K\} \in \mathbb{R}^{K \times D}$, which act as task-specific queries attending to relevant actor and scene regions. 
The overall task is to jointly predict: (1) individual action labels for all detected actors, (2) group membership assignments, and (3) group activity labels for the identified social groups.

\subsection{Feature Extractor}
We extract both global scene features and localized actor representations from each frame using a frozen VFM. To adapt the model for group activity understanding without full fine-tuning, we introduce PEFT-based conditioning to expose socially relevant semantics and a compact RoI-based actor token extraction strategy.

\noindent\textbf{PEFT-based Feature Extraction.} VFMs are pretrained on object-centric data \cite{Dino_multi_object}, making them less suitable for capturing social configurations such as interacting actors and group-level spatial arrangements without adaptation. We therefore employ PEFT-based conditioning (adapters and prompts) to expose these semantics from the frozen backbone. Group prompts consistently outperform adapters (Table~\ref{tab:vs_sota}), indicating more effective conditioning for multi-group scenarios; we thus adopt prompts as our primary mechanism. Prompts are learned per block but shared across all $T$ frames, providing consistent group-aware conditioning. We fix the number of prompts $K$ to the maximum number of annotated groups, enforcing a one-prompt-per-group design that encourages disentangled representations and improves coverage (Table~\ref{tab:number_group_tokens}). The resulting prompt-conditioned tokens capture group-aware semantics, from which actor tokens are extracted at the final layer via RoI pooling.

\noindent\textbf{RoI-based Actor Token Generation.} We extract actor‑specific tokens using 1 $\times$ 1 RoI pooling from the prompt-conditioned feature maps at the final backbone layer, producing a single embedding per actor. This compact representation preserves spatial precision and integrates seamlessly with transformer‑based reasoning. Compared to prior methods \cite{kim2024cafe,gad_vlm1} that use flattened 5 $\times$ 5 RoI grids, our design reduces token dimensionality without introducing additional spatial redundancy, enabling a leaner reasoning module.


\begin{figure}[ht]
	\centering
	\includegraphics[width=1\columnwidth]{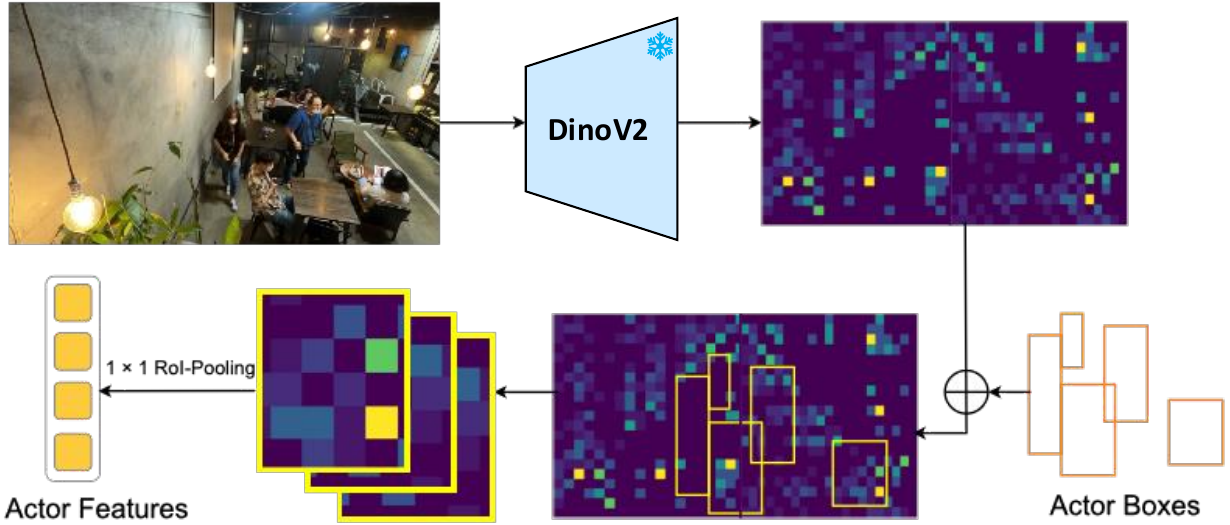}
	\caption{RoI-based actor token extraction pipeline. Given actor bounding boxes and feature maps from a frozen vision backbone, 1×1 RoI pooling is used to extract fixed-size actor tokens. These tokens serve as spatially grounded inputs for downstream reasoning in the ProGraD architecture. }
	\label{fig:cit}
    \vspace{-1em}
\end{figure}

\subsection{GroupContext Transformer}
The GroupContext Transformer (GCT) is a lightweight two‑layer decoder designed to reason jointly over actors, group tokens, and global scene context. Let $M$, $K$, $T$, and $D$ denote the number of actors, groups, frames, and feature dimensions, respectively; attention is applied by reshaping the tokens such that the temporal dimension $T$ is merged with the token dimension, enabling joint spatio-temporal attention.  The decoder operates in two stages: first, by inferring actor–group associations via the Grouping Attention Layer, then enriching tokens with global scene context via the Contextual Attention Layer. The outputs are enriched actor $\mathbf{A}_{c tx}$ and group tokens $\mathbf{G}_{ctx}$ used for predicting individual actions, group activities, and group membership scores.


\noindent\textbf{Grouping Attention Layer.} The first layer initializes a set of learnable group tokens $\mathbf{G}_{prompt}\in\mathbb{R}^{K\times D}$,  replicated across $T$ frames to form $\mathbf{G}_{init}\in\mathbb{R}^{K\times T\times D}$,
which serve as queries attending to the actor tokens $\mathbf{A}\in\mathbb{R}^{M\times T\times D}$  (keys and values). 
This cross‑attention aggregates information from all detected actors into the group tokens:
\begin{equation}
\mathbf{G}_{grp} = \text{LN} \left( \mathbf{G}_{init} + \text{FFN} \left( \text{Attn}(\mathbf{G}_{init}, \mathbf{A}, \mathbf{A}) \right) \right)
\label{eqn:grouping}
\end{equation}

Through this attention mechanism (Equation \ref{eqn:grouping}), group tokens aggregate information from potentially distant or temporally separated actors, forming a coherent representation of group composition. The subsequent residual connection, feed‑forward network (FFN), and layer normalization (LN) yield the updated group tokens $\mathbf{G}_{grp}$ that reflect actor-group interactions.

\noindent\textbf{Contextual Attention Layer.}
After learning group compositions, we enrich both actor and group representations by incorporating global scene context. We first concatenate the updated group tokens $\mathbf{G}_{grp}$ with the actor tokens $\mathbf{A}$ and define this joint set as  $Z = [\mathbf{A} ; \mathbf{G}_{grp}] \in \mathbb{R}^{(M + K) \times T \times D}$. These queries attend to the image features from the PEFT conditioned feature extractor $\mathbf{I} \in \mathbb{R}^{H_f W_f \times T \times D}$, which are scene-level patch tokens.

\begin{equation}
\mathbf{ [\mathbf{A}_{ctx}; \mathbf{G}_{\text{ctx}}]} = \text{LN} \left( Z + \text{FFN} \left( \text{Attn}(Z, \mathbf{I}, \mathbf{I}) \right) \right)
\label{eqn:contextual}
\end{equation}

This step injects contextual cues - such as spatial arrangement, background objects, and interaction-relevant regions - into the group and actor embeddings, enabling more accurate group activity detection. These enriched actor $\mathbf{A}_{ctx}$ and group  $\mathbf{G}_{ctx}$ embeddings are passed to the prediction heads for action recognition, group activity classification, and group membership inference.



\subsection{Prediction heads}
We use lightweight feed-forward networks for final predictions, following prior work \cite{kim2024cafe}. The group activity head classifies each group, the actor activity head predicts individual actions, and the membership head projects actor and group tokens into a shared embedding space. Actor–group affinities are computed via dot products, assigning each actor to the group with the highest score.


\subsection{Training Objective}
We adopt the multi-task loss from Practical GAD \cite{kim2024cafe}, combining: (i) individual action classification ($\mathcal{L}_{\mathrm{ind}}$), (ii) group activity classification ($\mathcal{L}_{\mathrm{group}}$), (iii) actor–group membership prediction ($\mathcal{L}_{\mathrm{mem}}$) via Hungarian matching \cite{kuhn1955hungarian}, and (iv) a contrastive group consistency loss ($\mathcal{L}_{\mathrm{con}}$) inspired by InfoNCE \cite{infoNCE} to enforce consistency among actors within the same group. 
The total loss is:

\begin{equation}
    \mathcal{L} = \mathcal{L}_{\text{ind}} + \sum_i \mathcal{L}_{\text{group}} + \lambda_m \sum_i \mathcal{L}_{\text{mem}} + \lambda_c \mathcal{L}_{\text{con}}
\end{equation}

Here, $\lambda_m$ and $\lambda_c$ are hyperparameters that control the relative importance of the membership prediction loss and the contrastive consistency loss, respectively.

\begin{table*}[ht]
  \centering
  \begin{tabular}{lccccccccc}
\toprule
\textbf{Models} & \textbf{Backbone} & \textbf{Full FT} 
& \multicolumn{3}{c}{\textbf{Split by Place}} 
& \multicolumn{3}{c}{\textbf{Split by View}} 
& \textbf{Trainable} \\
\cmidrule(lr){4-6} \cmidrule(lr){7-9}
& & 
& \textbf{Group} & \textbf{Group} & \textbf{Outlier} 
& \textbf{Group} & \textbf{Group} & \textbf{Outlier} 
& \textbf{Param(M)} \\
& & 
& \textbf{mAP$_{1.0}$} & \textbf{mAP$_{0.5}$} & \textbf{mIoU} 
& \textbf{mAP$_{1.0}$} & \textbf{mAP$_{0.5}$} & \textbf{mIoU} & \\
\midrule

ARG\cite{wu2019arg} & Inception-v3 & \ding{51} 
& 6.87& 28.44& 46.72&  11.03 & 34.50 & 56.61& $\sim$52\textsuperscript{*} \\

Joint\cite{ehsanpour2020joint} & I3D & \ding{51} 
 & 8.39 & 26.26 & 55.95& 14.05 &  36.08 & 60.09& $\sim$17\textsuperscript{*} \\

JRDB-base\cite{ehsanpour2022jrdb} & I3D & \ding{51} 
&  9.42 & 25.75 & 48.00& 15.43 & 34.81 & 60.43 & $\sim$16\textsuperscript{*} \\

HGC \cite{tamura2022huntinggroupcluestransformers} & I3D & \ding{51} 
& 3.47 & 18.46 & 52.56& 6.55 & 26.29 & 56.84 & $\sim$20\textsuperscript{*} \\

Practical GAD\cite{kim2024cafe} & ResNet-18 & \ding{51} 
& 10.85 & 30.90 & 63.84& 18.84 & 37.53 & 67.64& 22.77 \\

VLCMTN \cite{gad_vlm} & SAM & \ding{55} 
& 11.58 & 30.98 & 65.48& \textbf{19.67} & 41.06 & 68.29& $\sim$ 24\textsuperscript{*} \\

\midrule
\textbf{ProGraD}&&&&&&&&& \\
Decoder Only & DINOv2 & \ding{55} 
& 13.02& 32.68& 65.44& 15.88 & 41.58 & 68.55 
& \textbf{10.68} \\
Decoder + Adapter & DINOv2 & \ding{55} 
& \underline{16.18}& \underline{33.40}& \underline{66.94}& 16.17 & \underline{42.67} & \underline{68.56} 
& 11.87 \\
Decoder + Prompts & DINOv2 & \ding{55} 
& \textbf{17.03}& \textbf{39.11}& \textbf{67.89}& \underline{16.85} & \textbf{43.42} & \textbf{70.42}
& \underline{10.74} \\

\bottomrule
  \end{tabular}

  \caption{Comparison with state-of-the-art methods on the Café dataset under split-by-place and split-by-view settings. We report Group mAP at IoU thresholds 1.0 and 0.5, and Outlier mIoU. \textsuperscript{*}Approximate counts based on backbone and decoder sizes when not reported.}
  \vspace{-1em}
    \label{tab:vs_sota}
\end{table*}

%% file: sec/4_experiments.tex
\section{Experiments}
\noindent\textbf{Dataset and Implementation Details.} We evaluate ProGraD on two GAD benchmarks: Café \cite{kim2024cafe} and Social-CAD \cite{ehsanpour2020joint}. Café is a multi-camera dataset with 6-second clips from six café locations, featuring 3–14 actors per clip, annotated with 3.5M bounding boxes, group memberships, and six group activities: \textit{queuing}, \textit{ordering}, \textit{selfie}, \textit{eating}, \textit{studying}, and \textit{fighting}, plus \textit{singleton} labels for ungrouped individuals. Social-CAD contains 44 videos with bounding boxes, individual actions (\textit{walking}, \textit{talking}, \textit{queueing}, \textit{waiting}, \textit{crossing}), and group labels derived from the majority actor action.  Since Social-CAD primarily includes singleton or loosely connected actors, it serves as a benchmark for sparse interactions. 
We use DinoV2‑Base \cite{dinov2} (ViT‑B/14) as a frozen backbone, training randomly initialized group prompts with the two‑layer GroupContext Transformer and prediction heads are randomly initialized and trained end-to-end. We sample $T=5$ frames for Café and $T=1$ for Social-CAD, following the evaluation protocol of Practical GAD \cite{kim2024cafe}, as Social-CAD provides frame-level annotations where temporal aggregation is not required. Models are trained for 30 epochs using AdamW 
with a mini‑batch of 32. Loss weights are set to $\lambda_m=5.0$, $\lambda_c=2.0$, and temperature $\tau=0.2$. 


\noindent\textbf{Evaluation Metrics.} We follow the official evaluation protocols of both benchmarks. On Café, we report Group mAP \cite{kim2024cafe} at IoU 0.5 and 1.0, which jointly assess group activity classification and spatial localization, and Outlier mIoU, which evaluates identification of singleton actors. 
On Social-CAD \cite{ehsanpour2020joint} dataset, where group instances are predominantly singletons, we report Social Accuracy (percentage of correctly predicted social group assignments) and Membership Accuracy (proportion of actors correctly assigned to their social group, including singletons). Together, these metrics comprehensively capture classification, localization, and outlier detection performance.

\subsection{Analysis and Qualitative Results}

We compare ProGraD against prior state-of-the-art methods, including \textbf{clustering-based} approaches such as ARG \cite{wu2019arg}, Joint \cite{ehsanpour2020joint}, and  JRDB-Base \cite{ehsanpour2022jrdb} that rely on handcrafted features and clustering for group formation,  and \textbf{transformer-based} models like HGC \cite{tamura2022huntinggroupcluestransformers}, Practical GAD \cite{kim2024cafe}, and VLCMTN \cite{gad_vlm} that use attention-based decoders for group matching.

\noindent\textbf{Impact of VFM Backbone.} A natural question is whether ProGraD's gains stem simply from using a stronger backbone. To test this, we replace Practical GAD’s ResNet‑18 with DINOv2 (Table \ref{tab:backbone_comparison}). 
 This reveals a striking result: substituting the backbone alone actually \textit{degrades} performance - Group mAP$@$1.0 drops from 10.85 to 9.46. This suggests a mismatch between VFM representations and task-specific decoders designed for CNN features: expressive classifiers may learn signals not actually encoded in the backbone \cite{non_linear_dc_probes, non_linear_probing_classifer}, thereby distorting rather than leveraging their representational capacity. In contrast, ProGraD’s GCT is designed to align with frozen VFM features, preserving their quality while enabling structured actor–group reasoning. These results indicate that decoder–representation alignment, rather than feature quality alone, is a key factor in GAD, and highlight lightweight structured decoding as a promising direction for leveraging VFMs for group activity understanding.
 


\begin{table}[ht]
  \centering

{\fontsize{7}{9}\selectfont  
\setlength{\tabcolsep}{4pt}
  \begin{tabular}{lcccc}
\toprule
     \textbf{Models} &\textbf{Backbone}& \textbf{Group} & \textbf{Group} & \textbf{Outlier} \\
     &  &\textbf{mAP$_{1.0}$} & \textbf{mAP$_{0.5}$} & \textbf{mIoU} \\
    \midrule
    Practical GAD\cite{kim2024cafe} &ResNet-18 & 10.85 & 30.90 & 63.84 \\
    Practical GAD\cite{kim2024cafe} & DinoV2&9.46& 25.20& 65.04\\
    ProGraD (Ours, Frozen Backbone)             & DinoV2 &\textbf{17.03} & \textbf{39.11} & \textbf{67.89} \\
    ProGraD (Ours, Fully fine-tuned)             & DinoV2 &\textbf{20.42} & \textbf{44.14} & \textbf{67.71} \\
    \bottomrule
  \end{tabular}

    }
\vspace{-0.2em}
\caption{Backbone impact on Café. Replacing ResNet‑18 with DinoV2 in Practical GAD degrades performance, while ProGraD’s GroupContext Transformer fully exploits VFM features in both frozen and full‑FT modes, proving it a robust, scalable decoder.
}
\vspace{-0.5em}
\label{tab:backbone_comparison}
\end{table}


\noindent\textbf{Café dataset.} Table~\ref{tab:vs_sota} compares ProGraD with prior state-of-the-art on the Café benchmark under both Split-by-Place and Split-by-View settings. We evaluate three frozen backbone configurations that share the same GCT decoder but differ in PEFT conditioning: \textit{Decoder Only} (no conditioning), \textit{Decoder + Adapter} (lightweight adapter modules), and \textit{Decoder + Prompts} (learnable group prompts injected into the VFM input sequence). Unless stated otherwise, \textbf{ProGraD} refers to the Decoder + Prompts configuration.

On Split-by-Place, ProGraD achieves state-of-the-art performance with Group mAP@1.0 of 17.03, Group mAP@0.5 of 39.11, and Outlier mIoU of 67.89, outperforming Practical GAD by 6.2 points on Group mAP@1.0 and 8.2 points on Group mAP@0.5, while using less than half the trainable parameters (10.74M vs.\ 22.77M). Notably, even \textit{Decoder Only}, i.e., GCT without any PEFT conditioning, achieves 13.02 Group mAP@1.0, surpassing all existing methods. This demonstrates that GCT is a strong GAD decoder even without additional conditioning. \textit{Decoder + Adapter} further improves performance to 16.18 Group mAP@1.0, outperforming prior methods and showing that both PEFT strategies benefit from the structured relational decoder. However, ProGraD consistently outperforms \textit{Decoder + Adapter} across all metrics (17.03 vs.\ 16.18 on Group mAP@1.0 and 39.11 vs.\ 33.40 on Group mAP@0.5), suggesting that, in this setting, learnable group prompts provide more effective conditioning than adapters for complex multi-group scenarios.

On Split-by-View, ProGraD achieves Group mAP@1.0 of 16.85, Group mAP@0.5 of 43.42, and Outlier mIoU of 70.42, leading all methods on Group mAP@0.5 and Outlier mIoU. While VLCMTN \cite{gad_vlm} achieves a higher Group mAP@1.0 (19.67), it relies on a decoder similar to Practical GAD \cite{kim2024cafe} augmented with text-based knowledge from vision–language models. In contrast, ProGraD achieves competitive performance with a simpler decoder and fewer than half the trainable parameters. The narrower margin on Group mAP@1.0 compared to Split-by-Place is consistent with prior observations that frozen VFMs struggle with cross-view generalization, as viewpoint changes affect low-level spatial features that cannot be adapted without fine-tuning \cite{T-mask}. \textit{Decoder + Adapter} also performs strongly (16.17 Group mAP@1.0, 42.67 Group mAP@0.5), further indicating that GCT is the primary driver of performance gains across different PEFT conditioning strategies.

\begin{figure}[ht]
	\centering
	\includegraphics[width=1\columnwidth]{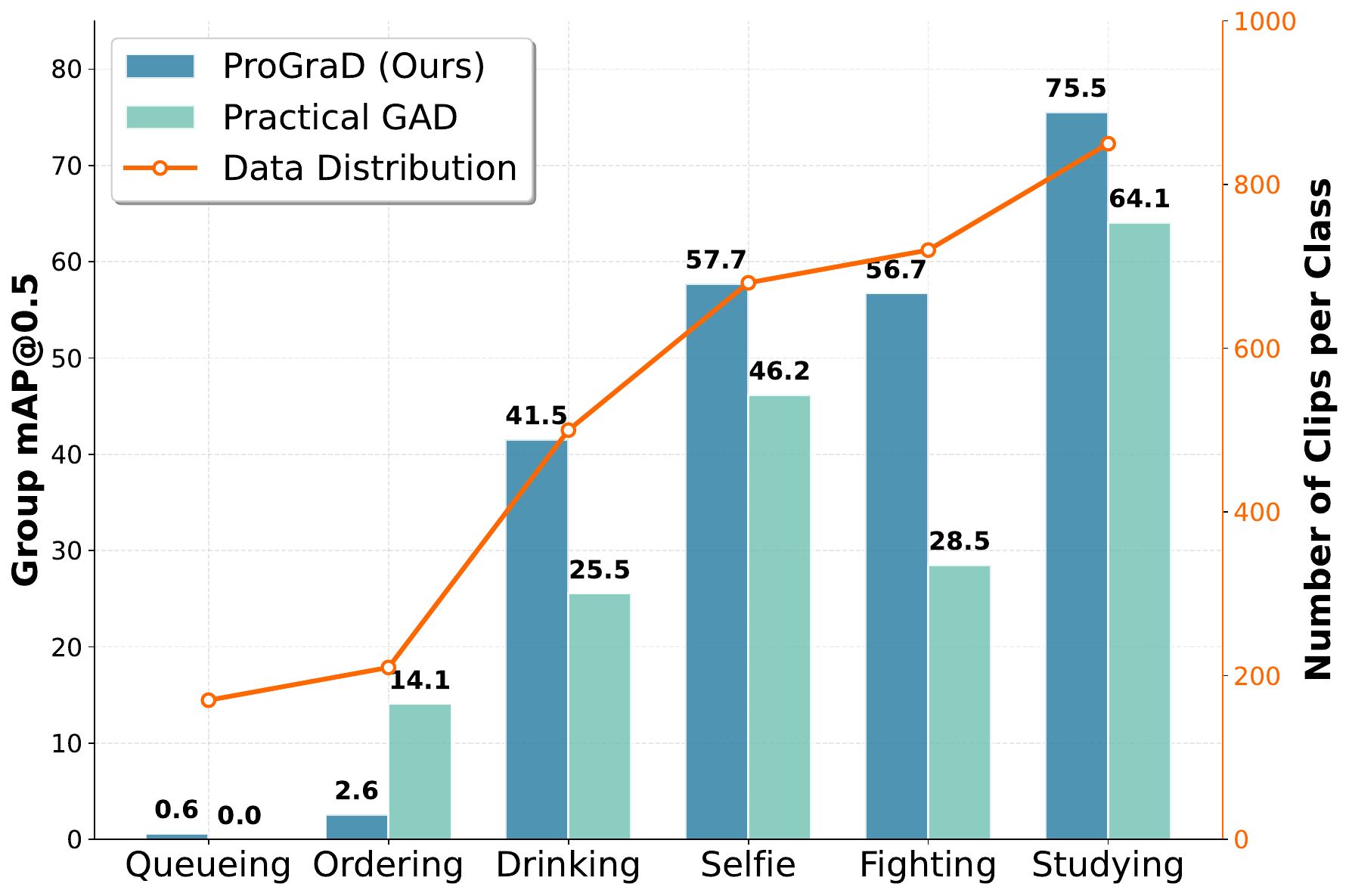}
	\caption{ Class-wise AP$@$0.5 comparison on the Café dataset. ProGraD outperforms Practical GAD on 5 of 6 classes.  Line overlay indicates ground-truth distribution of group activity classes. }
	\label{fig:classwise_dist}
    \vspace{-1em}
\end{figure}
\noindent\textbf{Class-wise Performance.} Figure~\ref{fig:classwise_dist} compares Group mAP$@$0.5 across activity categories with their frequency in the Café dataset. ProGraD delivers consistent gains over Practical GAD on most activities, Studying (75.5 vs.\ 64.1), Fighting (56.7 vs.\ 28.5), and Selfie (57.7 vs.\ 46.2), demonstrating the effectiveness of prompt-based group reasoning for visually distinctive activities. Performance drops on low-frequency classes such as Queueing and Ordering, underscoring the challenge of modeling rare group behaviors given the severe class imbalance in the dataset. This motivates future work on rebalancing strategies such as loss reweighting and few-shot prompt adaptation.

\noindent\textbf{Social CAD.} Table \ref{tab:sota_cad} compares ProGraD with prior methods on Social‑CAD.
ProGraD achieves a state‑of‑the‑art 70.00\% social accuracy and 88.33\% membership accuracy with only 10.74M trainable parameters, outperforming Practical GAD on social accuracy (69.20\%) while using less than half the parameters, and surpassing Joint \cite{ehsanpour2020joint} on membership accuracy (88.33\% vs.\ 83.00\%). The smaller performance gap compared to Café reflects the nature of Social-CAD - predominantly singleton or loosely connected actors leave limited room for improvement on group reasoning. Nonetheless, ProGraD's strong membership accuracy confirms its effectiveness in sparse interaction settings, demonstrating that prompt-guided adaptation generalizes well beyond dense multi-group scenarios. We do not report membership accuracy for Practical GAD as these results were unavailable and could not be reliably reproduced.


\begin{table}[ht]
  \centering

{\fontsize{8}{9}\selectfont  
\setlength{\tabcolsep}{4pt}
  \begin{tabular}{lcccc}
\toprule
     \textbf{Models} &\textbf{Backbone}&  \textbf{Social}&\textbf{Memb.}&\textbf{Trainable} \\
     &  & \textbf{Acc.}&\textbf{Acc.}&\textbf{Param(M)} \\
    \midrule
    ARG\cite{wu2019arg}&Inception-v3& 49.00 & 62.40 & $\sim$52         \\
    Joint\cite{ehsanpour2020joint}        & I3D & 69.00 &83.00&$\sim$17           \\
    Practical GAD\cite{kim2024cafe} & ResNet-18 & 69.20 &-&22.77  \\
    ProGraD (Ours)             & DinoV2 & \textbf{70.00}& \textbf{88.33}&\textbf{10.74} \\
    \bottomrule
  \end{tabular}

    }
      \caption{
      Comparison of Social‑CAD. ProGraD achieves SOTA social and membership accuracy with under half the parameters.
    \label{tab:sota_cad}
}
\vspace{-1em}
\end{table}

\subsection{Ablation Studies}

\noindent\textbf{Impact of GroupContext Transformer Components.}
To understand the contribution of each component in our GroupContext Transformer, we ablate the two key attention layers: the Grouping Attention Layer, responsible for modeling group membership, and the Contextual Attention Layer, which enriches both actor and group tokens with global scene context. As shown in Table \ref{tab:Two layer_ablation}, removing either component leads to a notable drop in performance. Without the Grouping Attention Layer, Group mAP$@$1.0 drops from 17.03 to 14.08, and Outlier mIoU drops to 64.56. Removing the Contextual Attention Layer causes a larger decline (10.97 mAP$@$1.0, 61.83 mIoU), underscoring complementary roles. Grouping Attention helps in group assignment for actors, whereas contextual attention captures broader relationships - both intra and inter-group relationships, providing critical context for accurate activity recognition.

\begin{table}[ht]
  \centering

{\fontsize{8}{9}\selectfont  
\setlength{\tabcolsep}{4pt}
  \begin{tabular}{lccc}
\toprule
     \textbf{Models} & \textbf{Group} & \textbf{Group} & \textbf{Outlier} \\
     &\textbf{mAP$_{1.0}$} & \textbf{mAP$_{0.5}$} & \textbf{mIoU} \\
    \midrule
    ProGraD wo Grouping Attn. Layer& 14.08 & 38.35 & 64.56\\
    ProGraD wo Contextual Attn. Layer&10.97	&30.20	&61.83\\
    ProGraD (Ours)             &\textbf{17.03} & \textbf{39.11} & \textbf{67.89} \\
    \bottomrule
  \end{tabular}

    }
          \caption{ Grouping aids actor-to-group assignment, while Contextual models global interactions for accurate recognition.
      }
            \label{tab:Two layer_ablation}
\end{table}

 \begin{table}[ht!]
  \centering

{\fontsize{9}{9}\selectfont  
\setlength{\tabcolsep}{4pt}
    \begin{tabular}{lccc}
      \toprule
      \textbf{Number of } & \textbf{Group} & \textbf{Group} & \textbf{Outlier} \\
      \textbf{Group Tokens} & \textbf{mAP$_{1.0}$} & \textbf{mAP$_{0.5}$} & \textbf{mIoU} \\
      \midrule
      4 & 12.81 & 35.30 & 64.95 \\
      \textbf{7 (Ours)} & \textbf{17.03} & \textbf{39.11} & \textbf{67.89} \\
      12 & 14.47 & 33.52 & 66.56 \\
      16 & 12.09 & 34.78 & 64.71 \\
      \bottomrule
    \end{tabular}

    }
      \caption{Effect of varying the number of group tokens. Using 7 tokens—matching the dataset’s group count—offers the best trade-off between expressivity and overfitting.
      }
          \label{tab:number_group_tokens}
          \vspace{-1em}
\end{table}

\noindent\textbf{Effect of Group Token Count.} We set the number of group prompts to 7, based on the maximum number of groups (including singletons) per video in the dataset, rather than treating it as a tunable hyperparameter. As shown in Table~\ref{tab:number_group_tokens}, this setting achieves the best performance (17.03 Group mAP@1.0, 39.11 mAP@0.5). Using fewer tokens ($K=4$) reduces coverage (12.81 mAP@1.0), while more tokens ($K=12,16$) introduce redundancy and overfitting, dropping to 14.47 and 12.09, respectively. This confirms that aligning token count with dataset structure is a principled design choice rather than a hyperparameter search.

\subsection{Qualitative Results}

\begin{figure}[ht!]
	\centering
	\includegraphics[width=1\columnwidth]{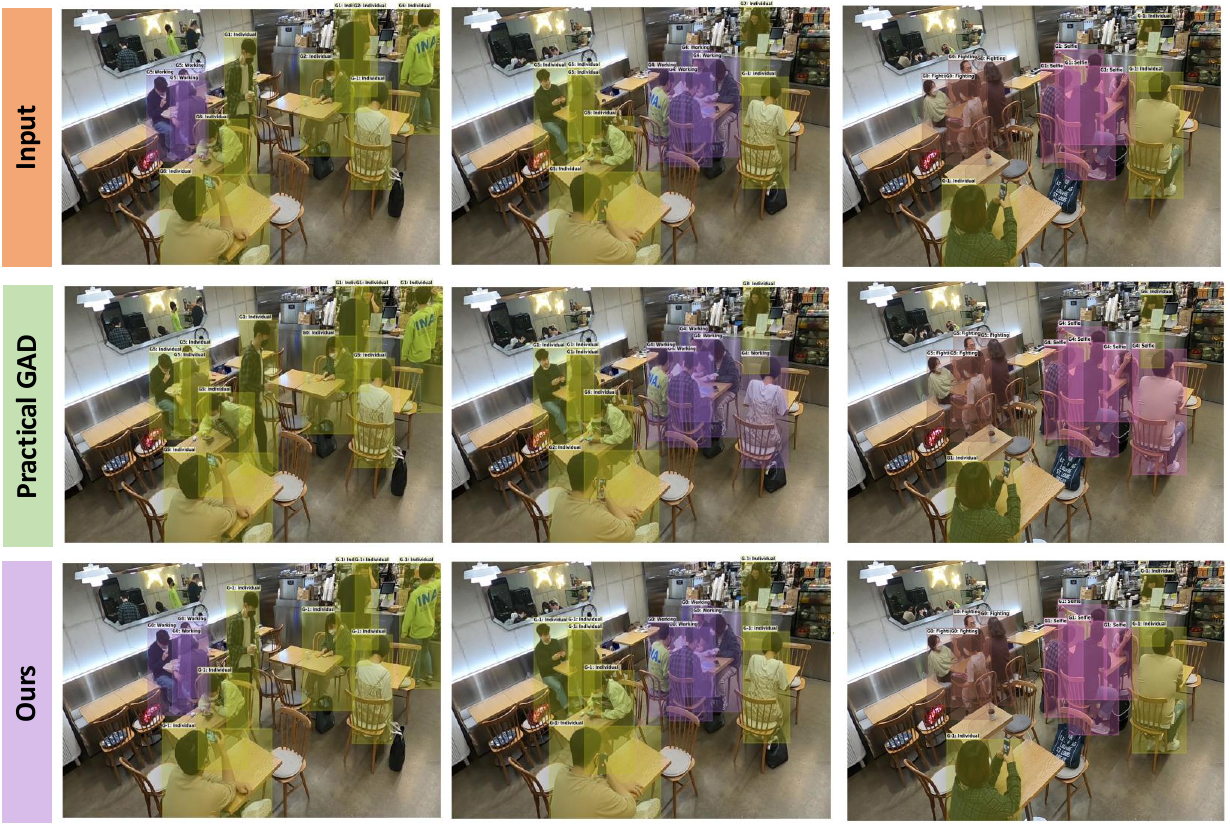}
	\caption{Qualitative comparison on the Cafè dataset. ProGraD produces cleaner group boundaries and more accurately separates closely positioned individuals from groups, demonstrating intent-driven rather than proximity-driven group reasoning.}
	\label{fig:quali_bb}
    \vspace{-1em}
\end{figure}

\noindent\textbf{Group Boundary Separation.} Figure \ref{fig:quali_bb} compares ProGraD with Practical GAD on group boundary separation, illustrating qualitative differences beyond the quantitative gains.
While Practical GAD tends to assign actors in close spatial proximity to a group, even when they are performing a different action, as members of that group, ProGraD correctly identifies them as outliers. This is evident across multiple examples: ProGraD accurately classifies proximate individuals as outliers and correctly identifies loosely interacting actors as a group, whereas Practical GAD fails on both counts. This reflects ProGraD's improved capability to model contextual and intent-driven group formations rather than relying solely on spatial proximity.

\begin{figure}[ht]
	\centering
	\includegraphics[width=1\columnwidth]{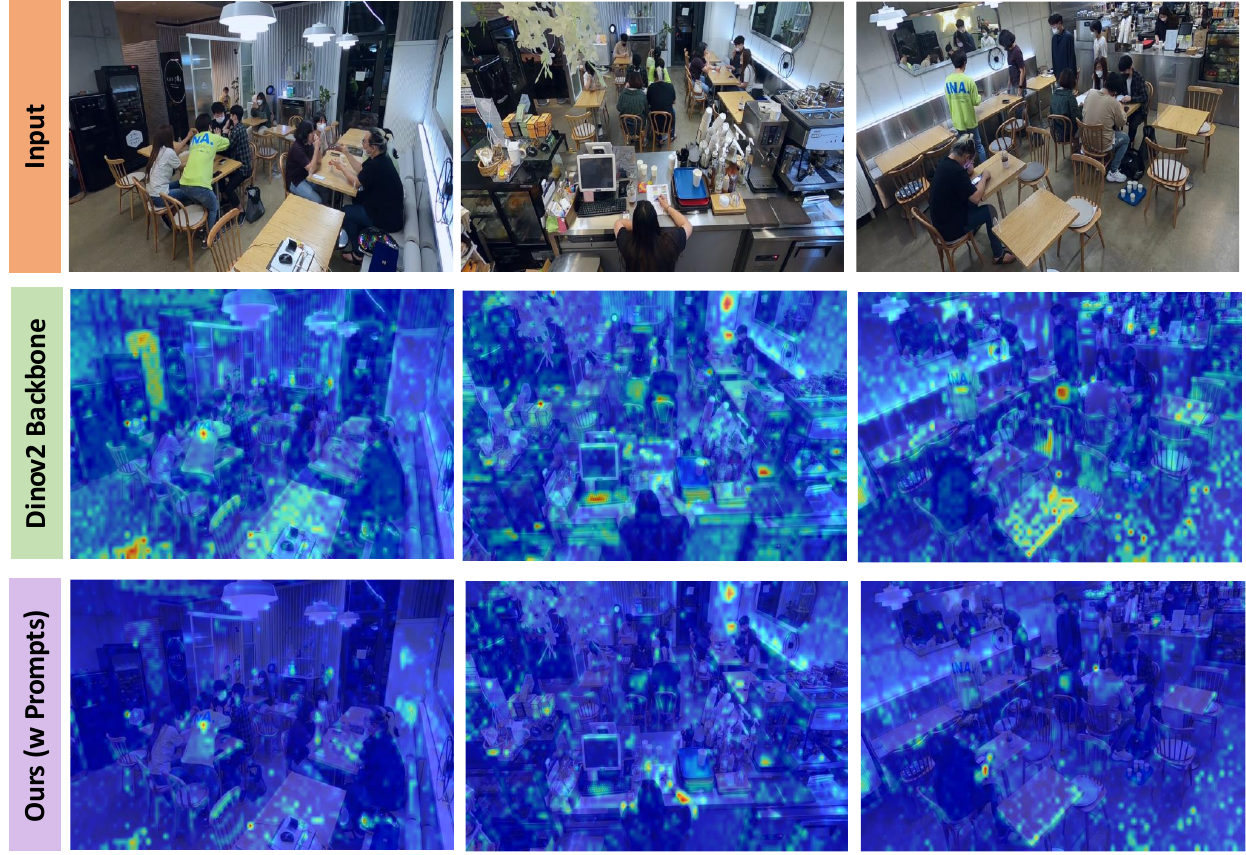}
	\caption{Attention map comparison before decoder. While naive DinoV2 attends broadly to objects and background regions, our prompt-guided backbone produces socially aware attention focused on groups and interaction-relevant areas.}
	\label{fig:attn_backbone}
    \vspace{-1em}
\end{figure}

\noindent\textbf{Effect of Prompt-Guided Backbone.} Figure \ref{fig:attn_backbone} compares attention maps from a naive DinoV2 backbone and our prompt‑guided backbone, both before the decoder, isolating the contribution of our PEFT-based conditioning. The naive backbone spreads its focus across objects and background areas, making it harder to extract useful group‑related features. In contrast, our prompt‑guided backbone focuses more clearly on people, their groups, and areas where interactions are likely (e.g., tables and shared spaces). This improved feature quality is consistent with the quantitative gains observed in our results, confirming the effectiveness of prompt-guided backbone adaptation for group reasoning.

\begin{figure*}[ht!]
	\centering
	\includegraphics[width=1\linewidth]{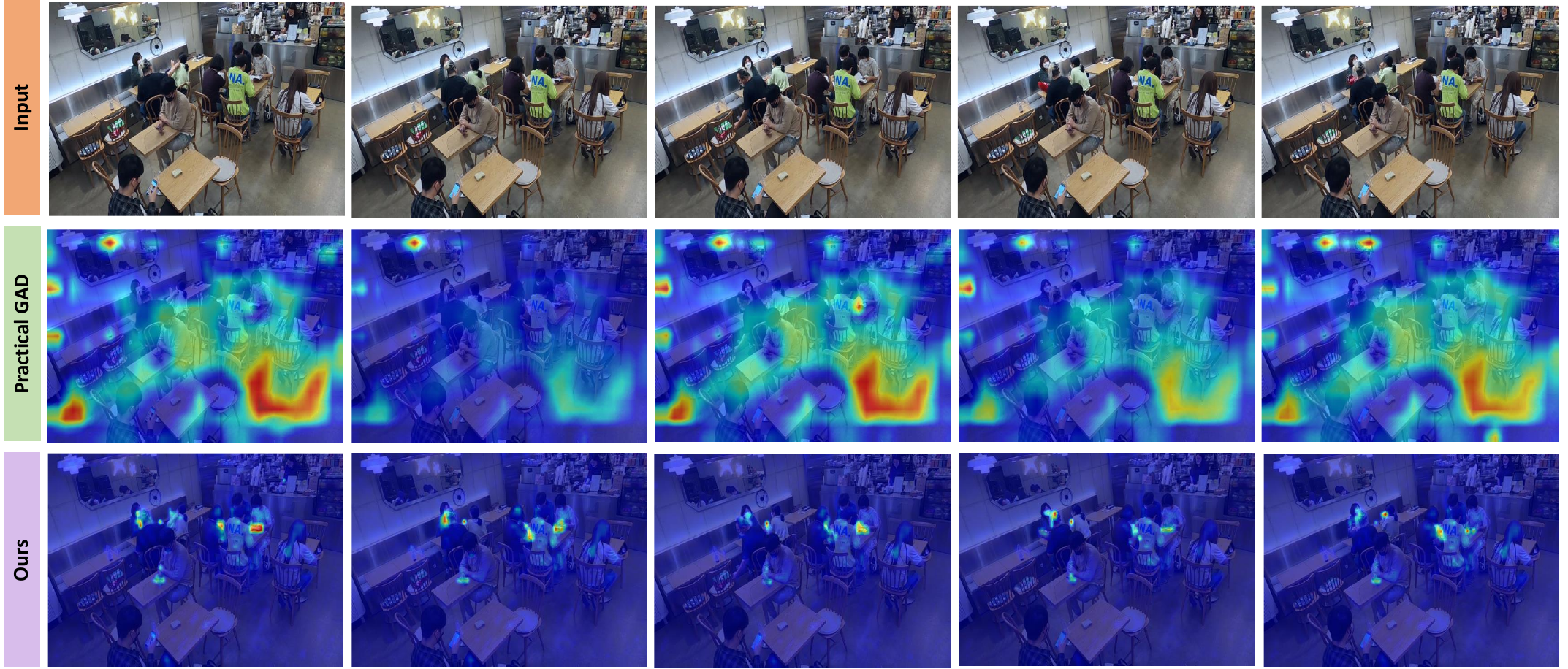}
	\caption{Attention map comparison. Across consecutive frames, ProGraD maintains focused attention on socially relevant regions, while Practical GAD shows more diffuse and inconsistent activation.}
	\label{fig:attn_temporal}
\end{figure*}

\noindent\textbf{Temporal Consistency.}
Figure \ref{fig:attn_temporal} illustrates attention heatmaps over consecutive frames from the same Cafè video, demonstrating ProGraD's temporal stability. ProGraD maintains consistent, sharply focused attention on socially relevant regions over time, including actors engaged in interactions and contextually important objects such as a book in a study group, hand movements in a fighting group, and a mobile phone held by an outlier. In contrast, Practical GAD exhibits broader, more diffuse activations extending into irrelevant background regions. This temporally stable, object-aware attention is consistent with ProGraD's superior group reasoning performance.

\begin{figure}[ht]
	\centering
	\includegraphics[width=1\columnwidth]{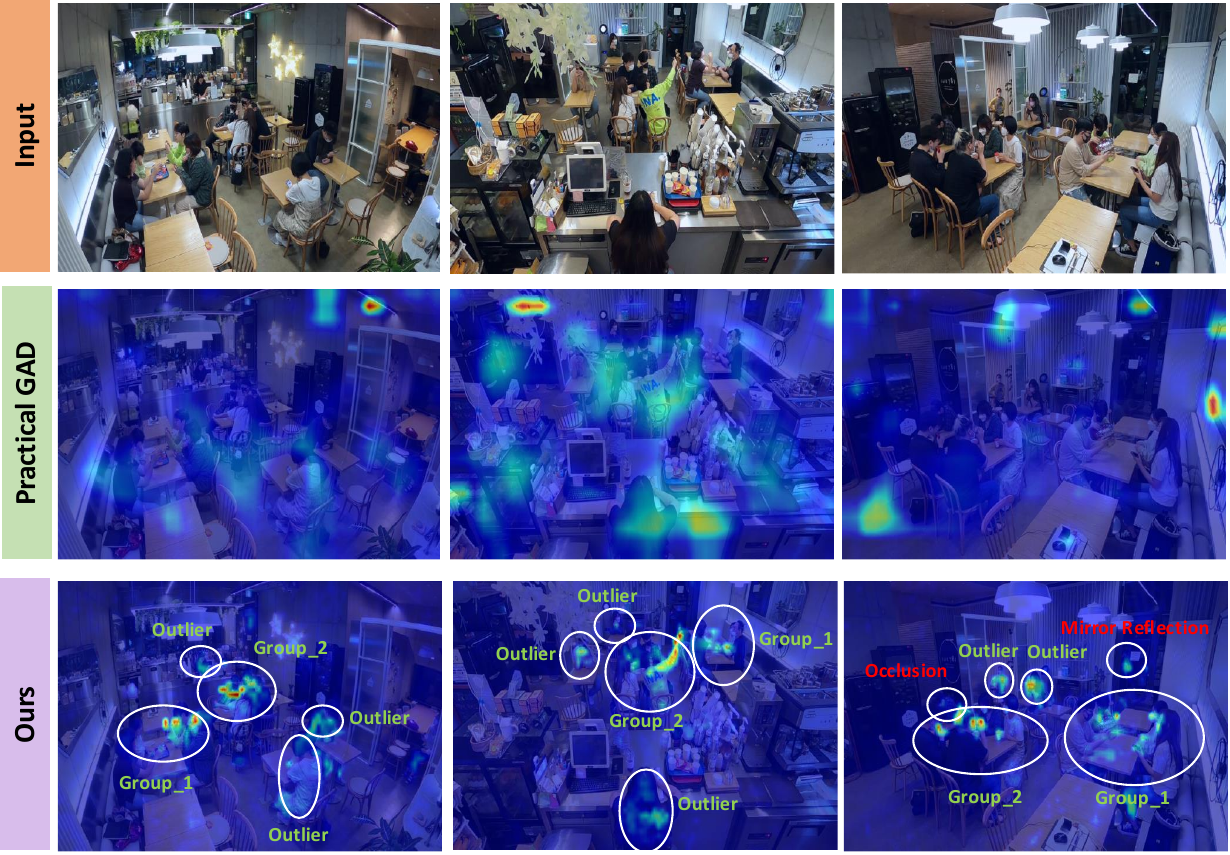}
	\caption{Attention map comparison across scenes. ProGraD produces sharper, clearly distinguishing social groups and outliers -enhancing interpretability. Red annotations mark failure modes (e.g., occlusion, mirror reflections), highlighting limitations and future challenges in group reasoning.}
	\label{fig:attn_sota}
\vspace{-1.5em}
\end{figure}

\noindent\textbf{Interpretability and Boundary Cases.}
Figure \ref{fig:attn_sota} presents a broader comparison of attention maps across diverse scenes, capturing both the strengths and limitations of ProGraD. Compared to Practical GAD's diffuse and overlapping patterns, ProGraD generates sharper and more structured attention maps that separate social groups and isolate outliers, resulting in more coherent group reasoning. Failure cases, marked in red, highlight challenges under visual ambiguity. In occluded scenes, obstructed spatial cues lead ProGraD to incorrectly include outliers in a nearby group. In environments with mirrors, the model mistakenly attends to reflections, disrupting group assignment. These examples motivate future extensions for spatial disambiguation and context-aware filtering.

%% file: sec/5_conclusion.tex
\section{Conclusion and Limitations}
We present \textbf{ProGraD}, a structured relational decoding framework that adapts frozen VFMs for Group Activity Detection via a lightweight \textbf{GroupContext Transformer (GCT)} and minimal PEFT-based conditioning.
A central finding of this work is that naively substituting VFMs for existing GAD decoders degrades performance, identifying decoder-representation alignment as the key bottleneck. ProGraD addresses this by separating actor-group formation from global context aggregation, achieving state-of-the-art results on Café and Social-CAD with only $\sim$10M trainable parameters in both frozen and fully fine-tuned settings. ProGraD also yields interpretable group-specific attention patterns, providing insight into actor-group reasoning for skilled activity assessment and feedback generation.


ProGraD has limitations that point to promising future directions. First, it assumes ground-truth actor detections; extending it to joint detection and grouping is an important direction. Second, the fixed prompt count $K$ (set to the dataset’s maximum group count) limits flexibility, motivating adaptive group token allocation for varying group sizes. Finally, while ProGraD is parameter-efficient, inference still requires a full forward pass through the frozen DINOv2 backbone.  Future work should explore more efficient strategies, such as feature reuse or adaptive token selection, to improve scalability.

\section*{Acknowledgments}
 The research published in this article is supported by the Deutsche
 Forschungsgemeinschaft (DFG) under Germany’s Excellence Strategy – EXC 2120/1. 
The authors also gratefully acknowledge the computing time provided on the high-performance computer HoreKa by the National High-Performance Computing Center at KIT. 
HoreKa is partly funded by the German Research Foundation (DFG).
